%% file: nicewebrl_neurips_2025.tex
\documentclass{article}

\newif\ifanonymous
\anonymousfalse  %

\newif\iflong
\longtrue %

\PassOptionsToPackage{numbers, compress}{natbib}
\ifanonymous
    \usepackage{neurips_2025}
\else
    \usepackage[preprint]{neurips_2025}
\fi

\usepackage[utf8]{inputenc} %
\usepackage[T1]{fontenc}    %
\usepackage{hyperref}       %
\usepackage{url}            %
\usepackage{booktabs}       %
\usepackage{amsfonts}       %
\usepackage{nicefrac}       %
\usepackage{microtype}      %
\usepackage{xcolor}         %
\usepackage{listings}

\usepackage{graphicx,amsmath,amsfonts,amssymb,fullpage,url,natbib}
\usepackage{algorithm}
\usepackage{algpseudocode}
\usepackage{array}
\usepackage{multicol}

\usepackage{parskip}
\usepackage{lipsum}
\usepackage{lipsum}  
\usepackage{wrapfig}

\usepackage{authblk}
\usepackage{multicol}

\usepackage{hyperref}

\definecolor{myred}{HTML}{E06666}
\definecolor{myblue}{HTML}{299DD3}
\definecolor{mygreen}{HTML}{159B27}
\definecolor{lightsteel}{HTML}{8f8f8c}
\definecolor{mediumsteel}{RGB}{82, 83, 94}
\definecolor{darksteel}{RGB}{54, 69, 79}
\definecolor{darkgreen}{RGB}{0,77,64}
\definecolor{prettycyan}{RGB}{0, 180, 180}  %
\definecolor{olive}{HTML}{9CB380}     %
\definecolor{orange}{HTML}{FECDAA}     %
\definecolor{salmon}{HTML}{FFA3A5}     %

\hypersetup{
    colorlinks=true,
    linkcolor=salmon,  %
    urlcolor=myblue,     %
    citecolor=prettycyan,  %
    filecolor=myblue     %
}

\definecolor{myorange}{HTML}{FF9B85}

\newcommand{\figref}[1]{{Figure~\ref{#1}}}

\newcommand{\envstage}{{\tt EnvStage}}
\newcommand{\nicewebrl}{{NiceWebRL}}

\title{\nicewebrl: a Python library for human subject experiments with reinforcement learning environments}

\author{%
    \begin{multicols}{2}
        \textbf{Wilka Carvalho}$^1$ \\
        \vspace{-8pt}
        $^1$Kempner Institute for the Study of Natural and Artificial Intelligence, Harvard University \\
        \texttt{wcarvalho@g.harvard.edu}
        \columnbreak

        \textbf{Vikram Goddla}$^{*\dagger2}$, \textbf{Ishaan Sinha}$^{*\dagger2}$, \\
        \textbf{Hoon Shin}$^2$ and \textbf{Kunal Jha}$^3$ \\
        $^2$Harvard College, $^3$University of Washington \\
        $^{*}$equal contribution, $^{\dagger}$core contributor
    \end{multicols}
}

\begin{document}
\maketitle

\begin{abstract}
We present NiceWebRL, a research tool that enables researchers to use machine reinforcement learning (RL) environments for online human subject experiments.
NiceWebRL is a Python library that allows any Jax-based environment to be transformed into an online interface, supporting both single-agent and multi-agent environments.
As such, NiceWebRL enables AI researchers to compare their algorithms to human performance, cognitive scientists to test ML algorithms as theories for human cognition, and multi-agent researchers to develop algorithms for human-AI collaboration.
We showcase NiceWebRL with 3 case studies that demonstrate its potential to help develop Human-like AI, Human-compatible AI, and Human-assistive AI.
In the first case study (Human-like AI), NiceWebRL enables the development of a novel RL model of cognition. Here, NiceWebRL facilitates testing this model against human participants in both a grid world and Craftax, a 2D Minecraft domain.
In our second case study (Human-compatible AI), NiceWebRL enables the development of a novel multi-agent RL algorithm that can generalize to human partners in the Overcooked domain.
Finally, in our third case study (Human-assistive AI), we show how NiceWebRL can allow researchers to study how an LLM can assist humans on complex tasks in XLand-Minigrid, an environment with millions of hierarchical tasks.
The library is available at \ifanonymous\url{https://anonymous.4open.science/r/nicewebrl-28BF}\else\url{https://github.com/KempnerInstitute/nicewebrl}\fi.
\end{abstract}

\section{Introduction}
\begin{figure}[!htp]
\centering
\includegraphics[width=1.0\textwidth]{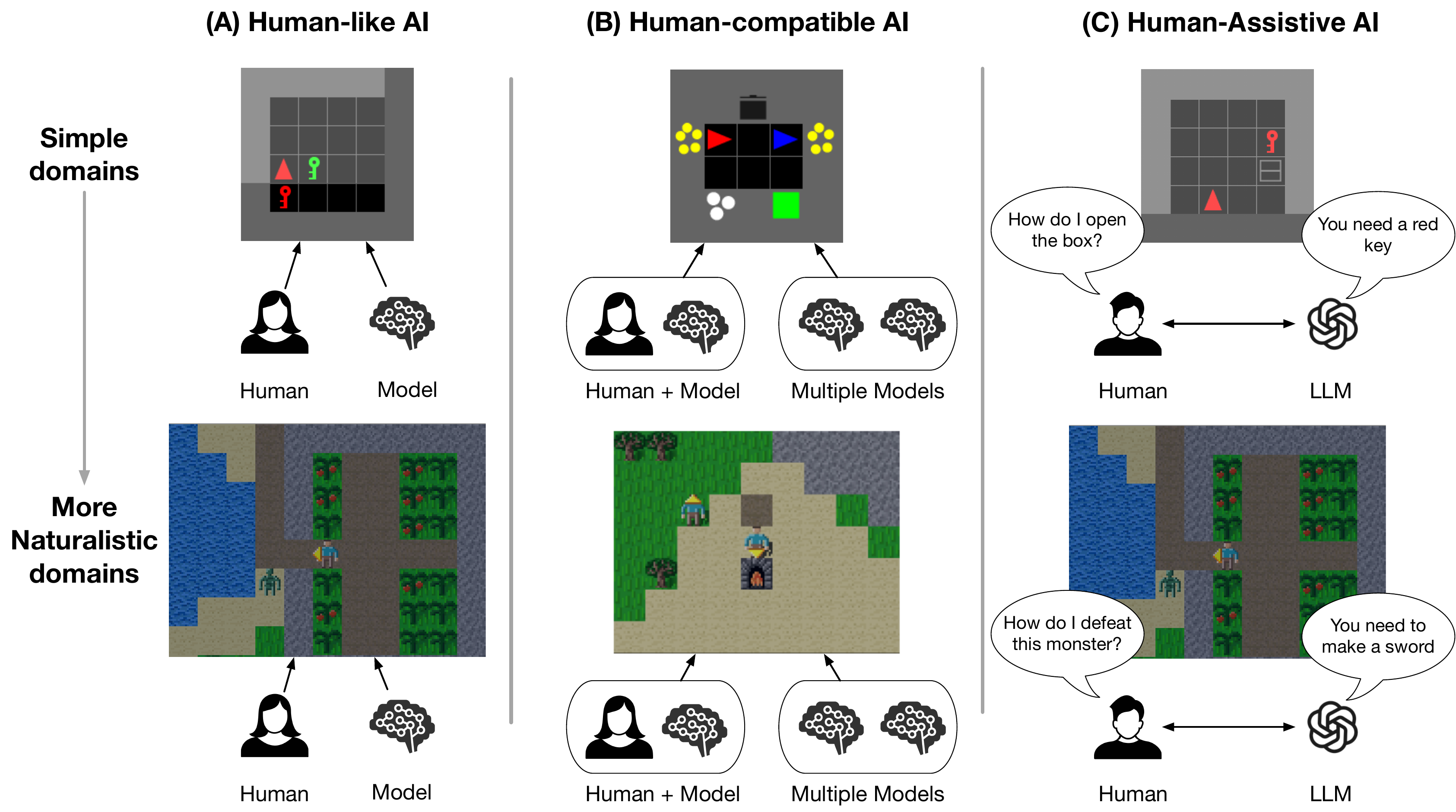}
\vspace{-20pt}
\caption{
    \textbf{\nicewebrl~is a meta-environment that enables the use of Jax-based environments to develop Human-like, Human-compatible, and Human-assistive AI.}
    (A) Researchers can compare how humans and AI complete tasks to evaluate if AI behaves in human-like ways.
    (B) They can study how AI coordinates with humans during task completion to assess if the AI has learned human-compatible social behaviors.
    (C) They can also integrate Large Language Models into their experiments to evaluate how effectively they combine prior knowledge with environmental perception to assist participants.
    Importantly, \nicewebrl~enables findings to generalize across potentially more complex domains.
    }
\label{fig:human-ai-comparisons}
\vspace{-10pt}
\end{figure}

\iflong
The last 20 years have seen a whirlwind of progress in Machine Learning (ML).
Reinforcement Learning (RL) agents have achieved superhuman performance on complex games such as Go~\citep{silver2017mastering};
computer vision systems can now process complex scenes~\citep{zhou2022learning,kirillov2023segment,radford2021learning};
and large language models (LLMs) increasingly act as our coding assistants and thought partners~\citep{achiam2023gpt,bubeck2023sparks}.

This progress motivates many researchers to study modern Artificial Intelligence (AI) agents in the context of human behavior.
\fi
Some ML researchers aim to improve AI systems by comparing them to humans, since humans can still provide an upper bound on the performance our systems can hope to achieve~\citep{mnih2015human,team2023human}.
For example, Minecraft~\citep{guss2019minerl,hafner2023mastering} remains a challenging exploration problem for machines but a fun exploration adventure for people~\citep{hjorth2021exploring,du2023can}.
In cognitive science, there is increased interest in asking whether these machines are human-like~\citep{lake2017building,ying2025benchmarking,carvalho2025naturalistic}.
Even if they are not, cognitive scientists are interested in using them as the basis for building human-like machines~\citep{lake2017building,carvalho2025preemptive}.
In multi-agent RL, many researchers are interested in whether these agents can act as adaptive partners to humans across the wide range of social settings they might be deployed in~\citep{carroll2019utility,russell2022human}.
This is increasingly relevant with LLMs, as they possess superhuman knowledge and are improving in their ``reasoning'' abilities~\citep{guo2025deepseek}.
A natural question is how well they can combine their prior knowledge with environment perception to assist us in completing complex tasks.
Collectively, these advances could have a wide impact---ranging from robotics~\citep{vemprala2024chatgpt} to education~\citep{holmes2022state} to healthcare~\citep{shaheen2021applications}.
Clearly, many are interested in building human-like, human-compatible, and human-assistive AI.

Despite this interest in human-centered AI development, pursuing research that integrates human subject experiments with modern ML libraries is currently a cumbersome process.
To run experiments with many participants, researchers leverage the internet to get large sample sizes~\citep{gureckis2016psiturk}.
Thus, most infrastructure is written in the web's programming language: JavaScript~\citep{finger2017labvanced,henninger2021lab,gureckis2016psiturk,de2015jspsych}.
Machine learning code, on the other hand, relies heavily on Python for model development~\citep{abadi2016tensorflow,paszke2019pytorch,jax2018github} and variants of C for developing fast environments for simulation~\citep{kolve2017ai2,ward2020using}.
Leveraging ML models or environments for human subject experiments currently requires setting up domain-specific server-client configurations that integrate Javascript, Python, and sometimes C.
Doing this for each domain makes the process even more cumbersome.

To address this challenge, we present \nicewebrl: a research tool that lets researchers leverage ML environments for human subject experiments (see \figref{fig:human-ai-comparisons}).
Integrating Python with JavaScript requires maintaining a connection between a remote Python-based server and a local Javascript-based client. This distance can cause latency issues when running online experiments.
To circumvent this challenge, \nicewebrl~exploits Jax~\citep{jax2018github}---a high-performance numerical computing library---to precompute and cache environment dynamics for arbitrary Jax-based environments.
\nicewebrl~then acts as a meta-environment for researchers to use arbitrary Jax-based environments in their human subject experiments.
Critically, \nicewebrl~allows researchers to program experiments entirely in Python by integrating with NiceGUI\footnote{\url{https://github.com/zauberzeug/nicegui/}}---a library that enables web developers to specify advanced Graphical User Interface (GUI) components entirely in Python.

\iflong
\textbf{Our contributions are as follows}.
(1) We present \nicewebrl,~a research tool that enables the use of Jax-based virtual environments for both developing artificial agents and for running human subject experiments (\S\ref{sec:nicewebrl}).
(2) We present 3 case studies that demonstrate how \nicewebrl~can support the development of Human-like AI~(\S\ref{sec:case-study-1}), Human-compatible AI~(\S\ref{sec:case-study-2}), and Human-assistive AI~(\S\ref{sec:case-study-3}).
(3) Our codebase, \ifanonymous\url{https://anonymous.4open.science/r/nicewebrl-28BF}\else\url{https://github.com/KempnerInstitute/nicewebrl}\fi, comes with several functional example folders using \nicewebrl~across these 3 scenarios. 
\fi

\iflong\input{setup}\fi

\iflong
\section{Case studies}
We present three case studies that display how \nicewebrl~can help in the development of human-like AI, human-compatible, and human-assistive AI.
In \S\ref{sec:case-study-1} and \S\ref{sec:case-study-2}, we'll highlight how \nicewebrl~contributed to new insights that span AI and cognitive science research on human-like and human-compatible AI models. We will first describe the experimental results of prior work and then discuss how this is made possible by \nicewebrl.
In \S\ref{sec:case-study-3}, we will present a proof-of-concept for how \nicewebrl~can support research on LLM-based human-assistive AI.
\fi

\subsection{Case study 1: Developing Human-like AI with \nicewebrl}\label{sec:case-study-1}
\begin{figure}[!htp]
  \centering
  \includegraphics[width=.9\textwidth]{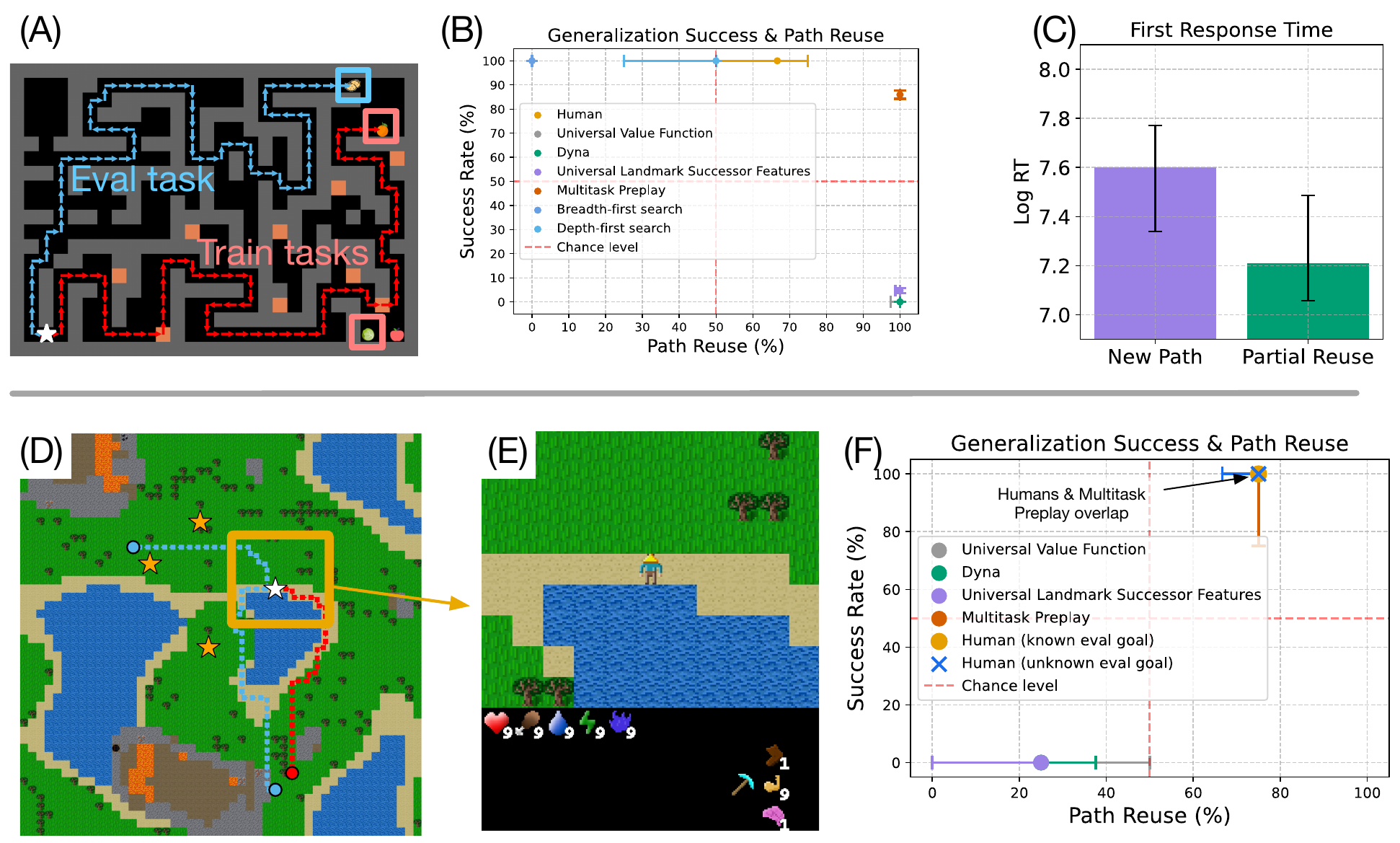}
  \caption{\textbf{Case study 1: \nicewebrl~enabled the development of a novel Deep RL cognitive model that generalizes to new tasks with the same qualitative behaviors as Humans across multiple domains}. (A, D): a gridworld and 2D minecraft environment that both Human participants and Deep RL models learned in. (B, F): behavioral results studying the same phenomena across the two domains. \nicewebrl~enabled developing a Deep RL cognitive model that could both (1) generalize to novel test goals in ways not permitted by previous methods, while (2) doing so with a similar suboptimal path reuse strategy that humans tend to exhibit. 
  Figures reproduced from~\citep{carvalho2025preemptive}.
  }
  \label{fig:case-study-1}
  \vspace{-10pt}
\end{figure}
\textbf{Developing a novel Deep RL cognitive science model with \nicewebrl~\citep{carvalho2025preemptive}}.
A central question in cognitive science is how people represent the environment to enable generalization to new tasks.
Successor features (SFs) are a mechanism for how an agent can cache expectations of what it will see when pursuing a policy~\citep{barreto2017successor}. 
Recent ML research has shown that SFs enable agents to repurpose policies for new tasks~\citep{barreto2018transfer,carvalho2023combining}.
Later, cognitive scientists showed that SFs also explained how people reuse prior policies for new tasks~\citep{tomov2021multi}.
However, the behavioral work was done in a small grid-world with $13$ states.
~\citet{carvalho2025preemptive} studied whether SFs could explain how humans reuse behaviors for new tasks in $2$ more complex domains: a maze gridworld (\figref{fig:case-study-1} A) and Craftax~\citep{matthews2024craftax}, a 2D minecraft domain (\figref{fig:case-study-1} D).
Across both domains, they set up training tasks where a test object was visible from along the optimal training paths (e.g. top-right corner of~\figref{fig:case-study-1} A). 
SFs could not generalize here. People could; however, when people reused a training path to a novel goal, their response times suggested that they were using a caching-based solution rather than something more flexible---but expensive---like planning at decision-time (\figref{fig:case-study-1} B-C).
They developed an algorithm termed Multitask Preplay which \textit{preemptively} learns solutions for unpursued tasks nearby training tasks by augmenting experience replay with small amounts of counterfactual simulation.
They found this algorithm both better accounted for the response times people exhibited and better predicted how they would reuse prior behaviors.
 These results generalized to Craftax, where participants and models had to navigate from partial observations of a large world with many objects (\figref{fig:case-study-1} E-F).

\textbf{Role of \nicewebrl.}
First, \nicewebrl~enabled comparing human behavior to advanced Deep RL algorithms including Successor Features. 
Second, \nicewebrl~enabled using the same infrastructure to study both Deep RL algorithms and human behavior in two domains of increasing complexity: a gridworld and Craftax.
This helped to ensure that findings were generalizable.
Finally, \nicewebrl~enabled the measurement of response times.
This helped to adjudicate between theories that predict more or less computation at decision-time.

\subsection{Case study 2: Developing Human-compatible AI with \nicewebrl}\label{sec:case-study-2}
\begin{figure}[!htp]
  \centering
  \includegraphics[width=1\textwidth]{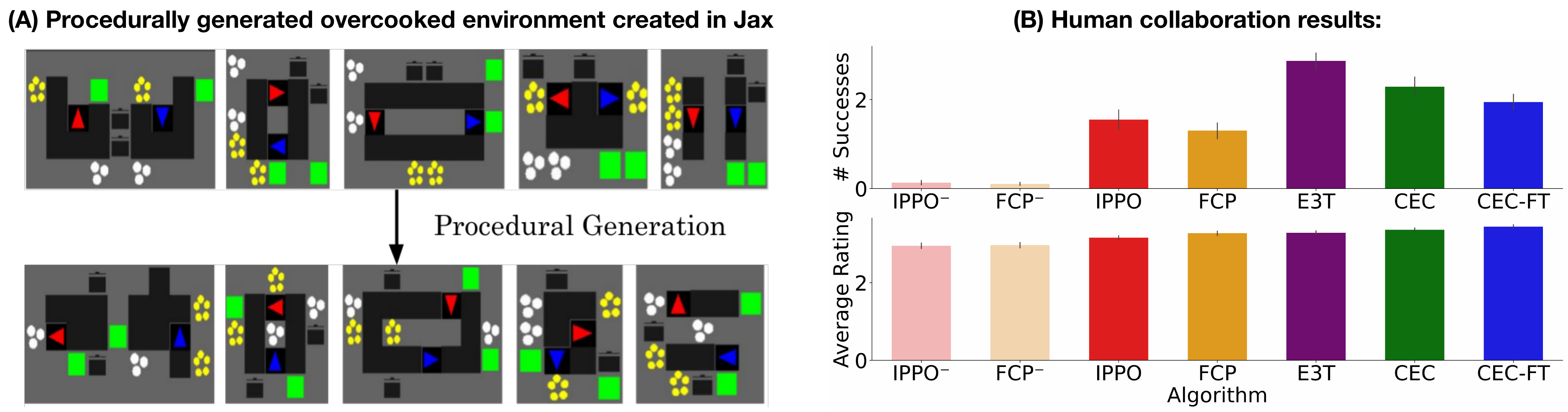}
  \vspace{-10pt}
  \caption{\textbf{Case study 2: \nicewebrl~enabled the development of a novel MARL algorithm that is more compatible with novel human partners.} (A) A procedurally-generated environment used to design a novel MARL algorithm: Cross-Environment Cooperation (CEC). Prior work had agents learn with diverse sets of agents. CEC has a \textit{single agent} play itself across millions of procedurally-generated environments. (B) While CEC \textit{succeeded less} than other methods when collaborating with humans (top), it succeeded in ways that were \textit{most favorable} to humans (bottom). Analysis suggested showed prior agents succeeded in less collaborative ways. 
        Figures reproduced from~\citep{jha2025cross}.
  }
  \label{fig:case-study-2}
\end{figure}
\textbf{Developing a novel Multi-agent reinforcement learning (MARL) algorithm for coordinating with humans using \nicewebrl~\citep{jha2025cross}}. 
One central question in MARL is how we can develop MARL agents that can generalize to human partners without human training data. 
One current benchmark for human-compatible AI is the  Overcooked domain~\citep{carroll2019utility} where agents must coordinate on basic cooking tasks.
The state-of-the art algorithm is ``Efficient End-to-End Training''~\citep[E3T;][]{yan2023efficient}, a ``Self Play'' algorithm that plays with---and tries to predict the actions of---a noisy variant of itself.
~\citet{jha2025cross} developed a novel algorithm, Cross-Environment Cooperation (CEC), where an agent plays only against itself but across millions of different procedurally-generated environments (\figref{fig:case-study-2} A).
They found that while E3T~\citep{yan2023efficient} was able to ``succeed'' on more episodes when collaborating with humans (\figref{fig:case-study-2} B), humans gave that agent a lower rating than CEC (\figref{fig:case-study-2} C).
The authors asked participants questions about their subjective experience using a Likert scale~\citep{likert1932technique} and found that CEC was rated as more ``adaptive'' and ``human-like''---despite succeeding \textit{less} than E3T. When the authors analyzed game trajectories, they found CEC would collide less with humans across environments.

\textbf{Role of \nicewebrl.}
First, \nicewebrl~enabled comparing multiple MARL algorithms in their ability to generalize to human partners.
Second, \nicewebrl~enabled the researchers to collect feedback from participants after every environment interaction stage using the ``Feedback'' stage object available in \nicewebrl~coupled with NiceGUI's data collection GUI elements.
This provided an easy way to get feedback from participants while agent-interaction data was fresh in their memories.
Third, \nicewebrl~stores all environment interactions so participant episodes could be analyzed post-hoc.
This enabled the researchers to analyze trajectories by participants and agents to determine what qualitative behaviors (such as colliding) were different between the different MARL algorithms.

\subsection{Case study 3: Developing Human-assistive AI with \nicewebrl}\label{sec:case-study-3}
\begin{figure}[!htp]
  \centering
  \includegraphics[width=.9\textwidth]{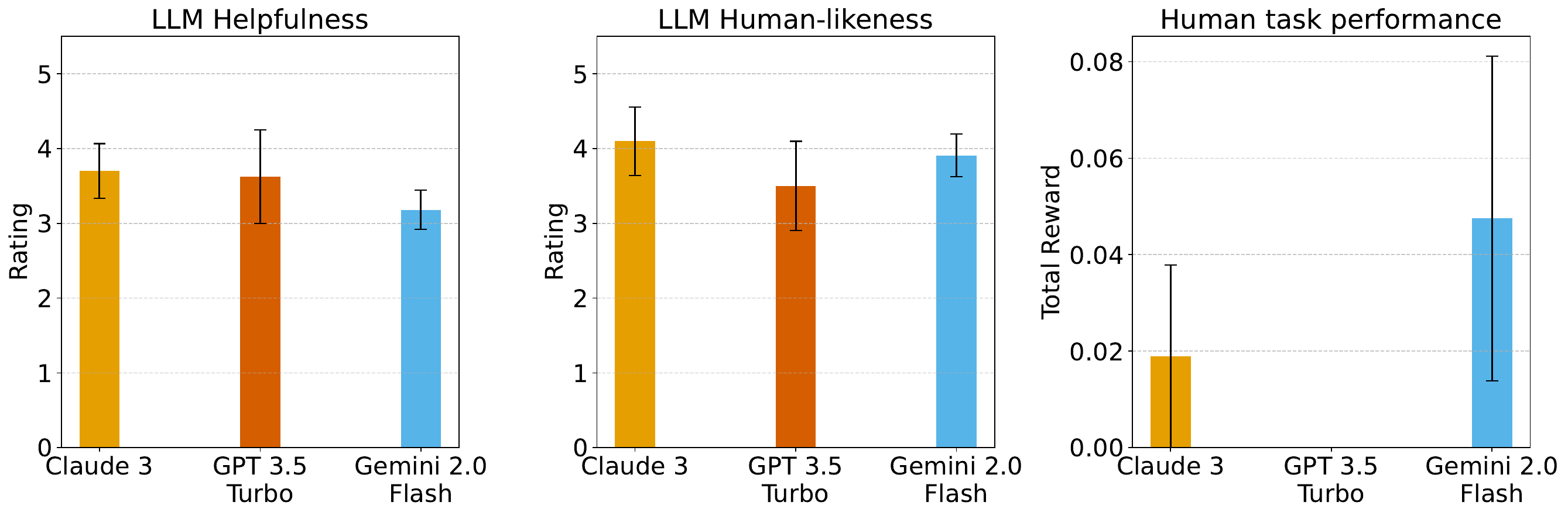}
  \caption{\textbf{Case study 3: Proof-of-concept experiment showing that \nicewebrl~enables comparing how different LLMs can assist people in completing tasks.}
    We had Claude 3 Opus, GPT 3.5 Turbo, or Gemini 2.0 Flash act as assistants for people completing tasks in the XLand-Minigrid domain~\citep{nikulin2024xland}.
    Each plot is showing the mean and standard error for $10$ subjects per model.
  }
  \label{fig:case-study-3}
\end{figure}
\textbf{Developing an LLM-assistant for sequential-decision making tasks in a virtual environment}.
We created a simple proof-of-concept experiment where people had to interact with tasks from the Xland-Minigrid domain~\citep{nikulin2024xland}.
To require assistance, they were given no task information but could ask an (anonymous) LLM assistant for help. 
\iflong
At the beginning of the experiment, users were randomly assigned either Claude 3 Opus, GPT 3.5 Turbo, or Gemini 2.0 Flash.
We set up our server so that it would interact with the LLMs via API calls\footnote{we provide an example of how to set up a web experiment with a local LLM in our examples folder}.
Importantly, the LLM assistants were given text descriptions of the ground truth environment state including information on:
    (1) the goal of the episode
    (2) the locations and identities of all objects in the environment
    (3) the rules of the environment (e.g. how objects interact when combined).
In principle, this can enable them to help users figure out the goal to maximize task reward.
In this proof-of-concept, each participant completed $3$ episodes, where a new task was sampled per episode.
After all $3$ episodes, participants answered two questions on a $5$-point scale, ``How helpful was the AI?'' and ``How human-like was the AI?''.
We collect data from $30$ participants via CloudResearch. We describe details around recruiting participants in \S\ref{appendix:human-details}. We show results in~\figref{fig:case-study-3}.
For this proof-of-concept, we used older models with cheap API calls. 
We don't expect that these results are representative of what is possible with frontier models.
\fi

\textbf{Role of \nicewebrl}.
\nicewebrl~enabled using an existing ML domain to develop an experiment that studied how an LLM could assist people on long-horizon tasks.
Additionally, the Feedback Stage object in \nicewebrl~enabled collecting feedback from participants about the LLMs at the end of the experiment. This could also be done after every episode or during episodes.

\iflong
\section{Discussion and conclusion}

We have presented \nicewebrl, a library for writing human subject experiments that leverage machine learning models and environments.
Importantly, by integrating with NiceGUI, experiments with sophisticated GUI components can be written entirely in Python.
NiceWebRL exploits Jax's compilation features and functional programming paradigm to reduce latency and enable multiple clients to interact with a single backend server.
We demonstrated the utility of \nicewebrl~with three case studies spanning both single-agent and multi-agent settings across $4$ domains: a custom gridworld, Craftax~\citep{matthews2024craftax}, Overcooked~\citep{flair2023jaxmarl}, and Xland-Minigrid~\citep{nikulin2024xland}.
In the first case study, we showed that \nicewebrl~could be used to measure human task performance and response times when developing a cognitive science model that could predict human behavior across two domains.
In the second case study, we showed that \nicewebrl~could be used to compare different MARL algorithms in their ability to generalize to humans without human training data.
Here, we showed that \nicewebrl's stage objects facilitated qualitative and quantitative analysis studying why different algorithms were rated to be better collaborators by human participants.
In our final case study, we showed that one could also use \nicewebrl~to run experiments that measure how well LLMs can assist people on sequential decision-making tasks under asymmetric information constraints.

\textbf{Limitations}. While we've demonstrated \nicewebrl's utility for human subject experiments, many improvement avenues remain.
We don't currently leverage Jax-based environment's ability to allow gradients to pass through their computational graph. Unsupervised environment design~\citep{dennis2020emergent} can exploit this to automatically generate environments with different properties. Like vision researchers use gradient descent to generate stimuli for humans~\citep{geirhos2018imagenet} and monkeys~\citep{wang2022tuning}, \nicewebrl~may be able to automatically generate environments for different target experimental conditions.
Another limitation is that \nicewebrl~currently only supports multi-agent domains with $2$ agents. Future work can look to design an $n$-dimensional generalizations of the \envstage object. 
Finally, while we precompute all next states to reduce latency, this may become prohibitive for large or continuous action spaces. Future work can integrate policy learning within the environment to select likely human actions for precomputing next states. 
Jax enables this policy to be incorporated into the environment's computational graph, which minimizes its computational cost.

\nicewebrl's reliance on Jax allows for a rich set of tools to improve future experiments.
Jax has a growing ecosystem of libraries spanning probabilistic programming~\cite{bingham2019pyro}, Bayesian inference~\citep{cabezas2024blackjax}, LLM development~\citep{geng2023easylm}, and general neural network development~\citep{flax2020github,kidger2021equinox}.
This enables researchers from various disciplines to leverage \nicewebrl~for human subject experiments regardless of their preferred modeling approach.
Thus, we are optimistic that \nicewebrl~can serve as a useful tool for future research developing Human-like, Human-compatible, and Human-assistive AI.
\fi

\clearpage
\bibliographystyle{plainnat}
\bibliography{mybib}

\clearpage
\appendix

\section{Formal domain description}\label{appendix:formal}

\newcommand{\Envstate}{S}
\newcommand{\envstate}{s}
\newcommand{\code}[1]{{\tt #1}}

Jax functions automatically compile to a fixed behavior when they receive their first input data.
As such, if one wants different  domain functionality across different \textit{contexts} (e.g. training vs. testing), the domain's functions typically need a ``{\tt env\_parameter}'' argument. 
Thus, Jax-based domains are naturally formulated as Partially Observable Contextual Markov Decision Processes (POCMDPs) $\mathcal{M}_{c}=\langle \Envstate, \mathcal{A}, \mathcal{X}, \mathcal{C}, \rho, P, R, O \rangle$~\citep{hallak2015contextual,kaelbling1998planning}.
Here, 
$\Envstate$ denotes the environment state space,
$\mathcal{A}$ denotes its action space,
$\mathcal{X}$ denotes (potentially partial) observations of the environment,
and $\mathcal{C}$ denotes a space of contexts that an MDP can be in.
\code{env\_parameter} then corresponds to an MDP's context $c \in \mathcal{C}$. 
It can be used to augment the initial state distribution $\rho_c(s_0)$ (e.g. having an agent start in different states in different contexts),
    the transition probabilities, $P_c(s'|s,a)$ (e.g. an agent's speed or strength can be changed in different contexts),
    the reward function $R_c(s)$ (e.g. different objects can be rewarded in different contexts),
    or the observation function $O_c(s)$ (e.g. objects can take on different colors in different contexts).

An episode proceeds as follows. An initial state $s_0 \in \Envstate$ is sampled from the initial state distribution $\rho_c(s_0)$.
When an agent takes an action $a\in\mathcal{A}$ in state $\envstate\in \Envstate$, the next state $\envstate'$ is sampled according to a next state distribution $\envstate' \sim P_c(\cdot|\envstate, a)$.
The agent then receives an observation $x'=O_c(\envstate')$ and reward $r'=R_c(o)$.
Note that $c$ is typically fixed within an episode.

\begin{figure}[!htp]
  \centering
  \setlength{\fboxsep}{10pt}  %
  \setlength{\fboxrule}{0.5pt}  %
  \fbox{%
    \begin{minipage}{\dimexpr\textwidth-2\fboxsep-2\fboxrule}   %
      \noindent\begin{minipage}[t]{0.48\textwidth}
        \centering\textbf{Server-side Operations}
      \end{minipage}%
      \hfill%
      \begin{minipage}[t]{0.48\textwidth}
        \centering\textbf{Client-side Operations}
      \end{minipage}
      \begin{minipage}[t]{0.48\textwidth}
        \begin{algorithmic}[1]
        \Statex Input: env context parameters $c$
        \Statex
        \Statex At time $t = 0:$
        \State $s_0, o_0 = \text{env.reset(c)}$
        \State $\{(s_1, o_1) = \text{env.step}(s_0, a, c)\}_{a \in \mathcal{A}}$
        \State cache $s_{\tt next} = \{s_1\}$
        \State send $o_0$ and $o_{\tt next} = \{o_1\}$ to the client
        \State
        \State
        \State
        \State
        \State 
        \State 
        \Statex
        \Statex At time $t = 1, 2, \ldots:$
        \State receive and store $(a_{t-1}, t_1, t_2)$
        \State select $s_t \in s_{\tt next}$ corresponding to $a_t$
        \State $\{s_{t+1}, o_{t+1} = \text{env.step}(s_t, a, c)\}_{a \in \mathcal{A}}$
        \State send $o_{\tt next} = \{o_{t+1}\}$ to the client
        \State update $s_{\tt next} = \{s_{t+1}\}$
        \end{algorithmic}
      \end{minipage}%
      \hfill%
      \begin{minipage}[t]{0.48\textwidth}
        \begin{algorithmic}[1]
        \Statex
        \Statex
        \Statex
        \State 
        \State 
        \State 
        \State display $o_0$ and record time $t_1$
        \State cache $o_{\tt next}$
        \State participant selects action  $a$
        \State record time $t_2$
        \State send $a_0$, $t_1$, $t_2$ to the server
        \State select $o_1 \in o_{\tt next}$ corresponding to $a_0$
        \State display $o_1$ and record $t_1$
        \Statex
        \Statex
        \State
        \State 
        \State
        \State cache $o_{\tt next}$
        \State participant selects $a_t$
        \State record time $t_2$
        \State send $a_t$, $t_1$, $t_2$ to the server
        \State select $o_{t+1} \in o_{\tt next}$ corresponding to $a_t$
        \State display $o_{t+1}$ and record $t_1$
        \end{algorithmic}
      \end{minipage}
    \end{minipage}
  }
  \caption{\textbf{Server-client Human-Environment Interaction Protocol.} Note that we omit displaying reward $r$ due to space constraints.}
  \label{alg:server_client}
\end{figure}

Let \code{env} be the programmatic object representing a domain.
In our library (and many RL libraries), $s_0, o_0=\code{env.step}(c)$ essentially plays the role of sampling from the initial state distribution and computing the corresponding observation for the agent.
The standard practice is to have $s_{t+1}, o_{t+1}, r_{t+1} = \code{env.step}(s_t, a_t, c)$ implement (a) sampling a new state (b) computing the corresponding reward, (c) computing the observation that an agent will get.

\section{Descriptions of stage types}\label{appendix:stage-types}
\begin{figure}[!htp]
    \centering
\includegraphics[width=0.9\textwidth]{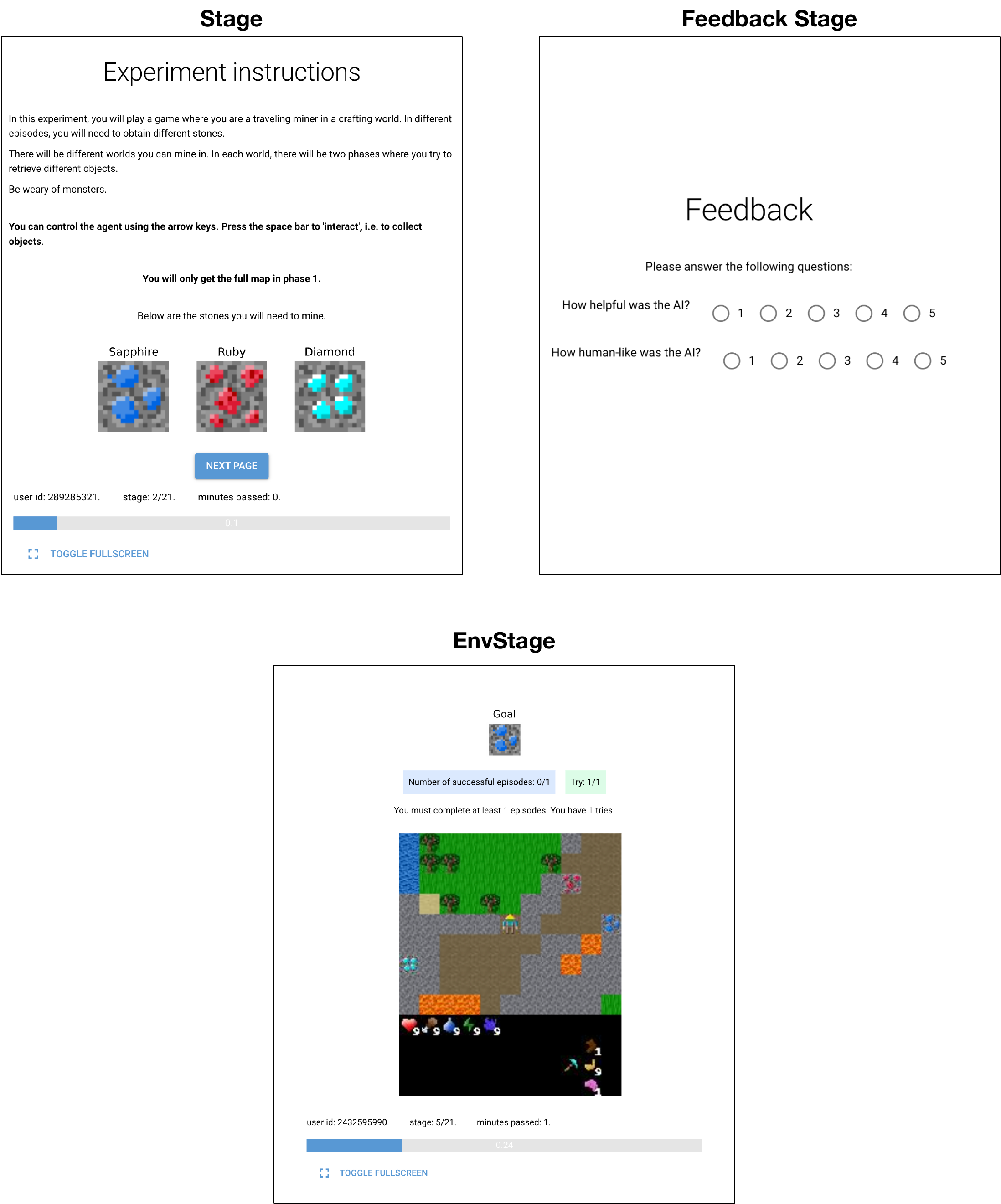}
\caption{\textbf{Examples of different kinds of stages.}}
\label{fig:stage-examples}
\end{figure}

Currently, there are three basic stage classes, though more can easily be added. 
\begin{enumerate}
    \item \code{Stage}: used to display instructions or information to a participant.
    \item \code{FeedbackStage}: used to collect information from participants. Typically involves an interactive screen that does \textit{not} interact with the environment.
    \item \code{EnvStage}: used to interact with an environment. It takes as input an environment and environment parameters. We describe how \nicewebrl~uses this abstraction to have a remote server-side program display images to one's local web-browser client in~\figref{alg:server_client}.
\end{enumerate}
We present examples of each in~\figref{fig:stage-examples}.

\section{Computing resources}\label{appendix:computing}

For details on case study 1 or 2, please see~\citep{carvalho2025preemptive} or \citep{jha2025cross}, respectively.
For case study 3, 
experiments were conducted using computing infrastructure from the \url{fly.io} platform with the ``performance-2x'' configuration. This is a machine with 4GB of RAM. The machine had no GPU. Even in this setting, Jax's compilation features provide a significant speed up to environment computation.

\section{Human subject experiment details}\label{appendix:human-details}
Our study is approved by the \ifanonymous University\else Harvard University\fi~IRB.
All subjects were recruited with \url{https://www.cloudresearch.com/} and provided informed consent. We provide the consent form in the GitHub example. Participants were compensated $\$4$ for completing the task.  The average task completion time was 23.33 minutes.
At the beginning of each experiment, the participants provided demographic information (age and gender, coded as male or female).

\end{document}

%% file: setup.tex
\section{Related Work}
\nicewebrl~is a meta-environment for leveraging Jax-based virtual environments in online human subject experiments.
It facilitates the development of Human-like AI, Human-compatible AI, and Human-assistive AI. 
There is a rich literature on these topics.
Researchers have created desiderata for measuring how ``Human-like'' AI agents are~\citep{ying2025benchmarking,ying2025benchmarking};
    they have developed benchmarks that test for ``human-like'' abilities~\citep{zhou2023far,sclar2023minding};
    and books~\citep{russell2022human} and articles~\citep{collins2024building} have been written about how to enable human-compatible AI.
There is a rich literature on how general AIs can assist humans~\citep{hadfield2016cooperative,shah2020benefits} and a growing literature on how LLMs can assist humans on tasks~\citep{liu2023agentbench,wu2023smartplay}.
Our focus is on enabling researchers to run human subject experiments with modern ML models and environments. Below we review the most relevant literature.

\textbf{Running human subject experiments}. JavaScript is the only programming language that can be run on modern web browsers. As a consequence, most tools for human subject experiments target JavaScript-based development.
For example, Labvanced~\citep{finger2017labvanced}, lab.js~\citep{henninger2021lab}, and Psiturk~\citep{gureckis2016psiturk} all provide GUI interfaces for designing JavaScript-based web experiments.
While there are Python-based tools for developing human subject experiments like Psychopy~\citep{peirce2007psychopy}, they only permit local experiments and an accompanying JavaScript library must be used for writing web experiments.
JsPsych~\citep{de2015jspsych} is a JavaScript library that facilitates programmatic definitions of experiments like \nicewebrl.
While each are useful, none provide utility for leveraging modern ML models and environments (written in Python) for online human subject experiments. This is precisely the gap that \nicewebrl~aims to fill.

\textbf{Comparing natural intelligence and artificial intelligence in virtual environments}. 
Many of the efforts that compare natural and artificial intelligence rely on the Unity game engine, which allows for the development of 3D environments with realistic physics and sensory observations~\citep{ward2020using}.
Cobel-RL developed a 3D environment for studying neuroscience questions around spatial navigation with Deep learning based RL models (i.e. Deep RL models)~\citep{diekmann2023cobel}.
The Animal-AI environment was developed to study whether Deep RL models displayed cognitive abilities associated with animals~\citep{beyret2019animal}.
Most similar to \nicewebrl~is PsychLab which explicitly aims to compare Humans to Deep RL models across several classic experimental paradigms like visual search, multiple object tracking, and random dot motion discrimination~\citep{leibo2018psychlab}.
However, the following challenges its adoption.
(1) Its API is in Lua, which limits accessibility, whereas \nicewebrl~is purely in Python, (2) it lacks precise response time measurements which \nicewebrl~provides, (3) its reliance on Unity makes it challenging to run online experiments. While \nicewebrl~relies on Python, it is able to run online but maintain fast environment performance thanks to its reliance on Jax.

\section{Background}
\begin{wrapfigure}[19]{r}{0.5\textwidth}
  \vspace{-20pt}
  \centering
  \includegraphics[width=0.48\textwidth]{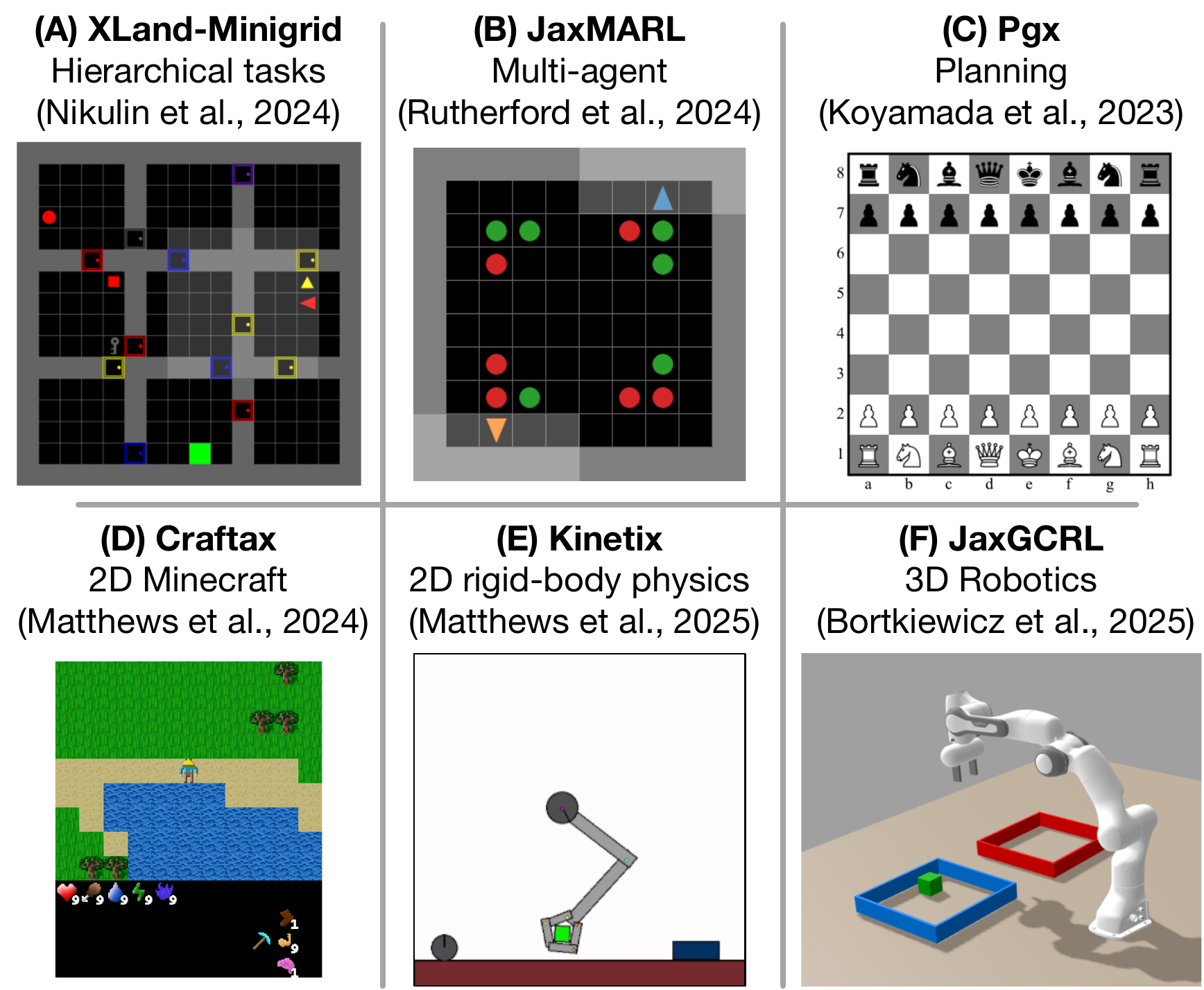}
  \caption{\textbf{Examples of Jax-based environments in the Jax ecosystem.} 
  All of these can be leveraged with \nicewebrl~to develop Human-like, Human-compatible, or Human-assistive AI.}
  \label{fig:environments}
\end{wrapfigure}
\textbf{The Jax ecosystem}.
Jax~\citep{jax2018github} is a Python library for high-performance numerical computing and machine learning with NumPy-like syntax. 
It follows a functional programming paradigm where functions are stateless, only defining input transformations. 
It achieves fast computation through JIT compilation and includes tools for tracking random number generators, enhancing reproducibility.
Most relevant to \nicewebrl, there is a growing set of environments including XLand-Minigrid, which has millions of hierarchical tasks for studying long-horizon generalization~\citep{nikulin2024xland}; JaxMARL, which has multi-agent environments for studying coordination~\citep{flair2023jaxmarl}; Pgx, which has planning environments for studying reasoning~\citep{koyamada2023pgx}; Craftax, a large procedurally-generated open-world environment that enables studying exploration and generalization~\citep{matthews2024craftax}; Kinetix, a procedurally-generated physics-based environment for studying physical reasoning~\citep{matthews2024kinetix}; and JaxGCRL, a goal-conditioned robotics environment for studying 3D manipulation tasks~\citep{bortkiewicz2025accelerating}. \nicewebrl~can be used with all of these environments.

\begin{wrapfigure}[19]{r}{0.5\textwidth}
  \centering
  \vspace{-15pt}
  \includegraphics[width=0.48\textwidth]{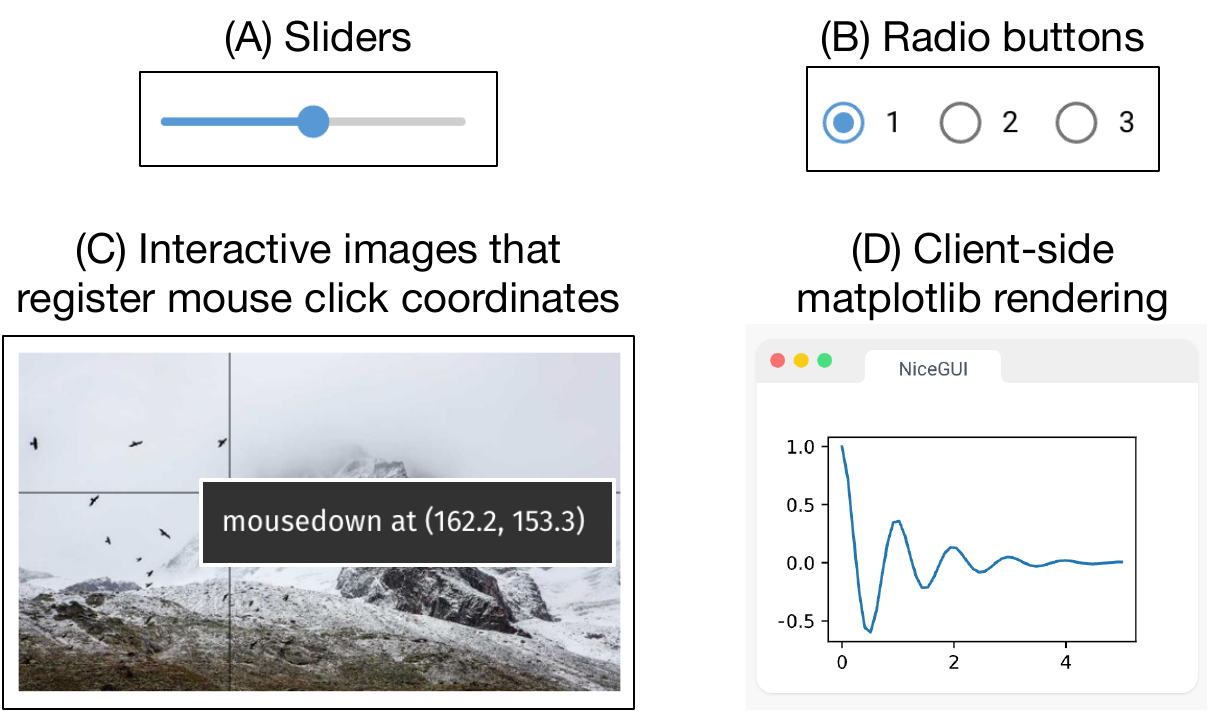}
  \caption{\textbf{Examples of advanced GUI capabilities provided by NiceGUI that can be leveraged with \nicewebrl.} Researchers can create (A) sliders for reporting numerical scores (B) radio buttons for selecting choices. (C) Beyond recording key strokes, researchers can record actions as $(x,y)$-coordinate selections on images or (D) leverage familiar plotting tools such as matplotlib for presenting graphs to participants.}
  \label{fig:nicegui}
\end{wrapfigure}
\textbf{Python-based web development} requires maintaining a Python-based web server that communicates with clients that operate in JavaScript.
We build on NiceGUI which comes with tools for handling many concurrent client connections asynchronously.
When sending large data packets between a client and a server, web socket connections are needed for real-time communication. 
NiceGUI's web socket implementation facilitates setting up persistent connections by having web sockets automatically reopen when connections close unexpectedly. 
This is key to having seamless human experiments with Python-based environment backends.
Finally, and equally important for online experiments, NiceGUI enables researchers to use Python to build responsive GUI components without any JavaScript knowledge. 
We provide examples of GUI components provided by NiceGUI in~\figref{fig:nicegui}.
All of these can be used with \nicewebrl.

\section{\nicewebrl}\label{sec:nicewebrl}
\nicewebrl~is a Python library that leverages Jax and NiceGUI to create online interfaces for 
\begin{wrapfigure}[28]{r}{0.65\textwidth}
    \centering
    \includegraphics[width=\linewidth]{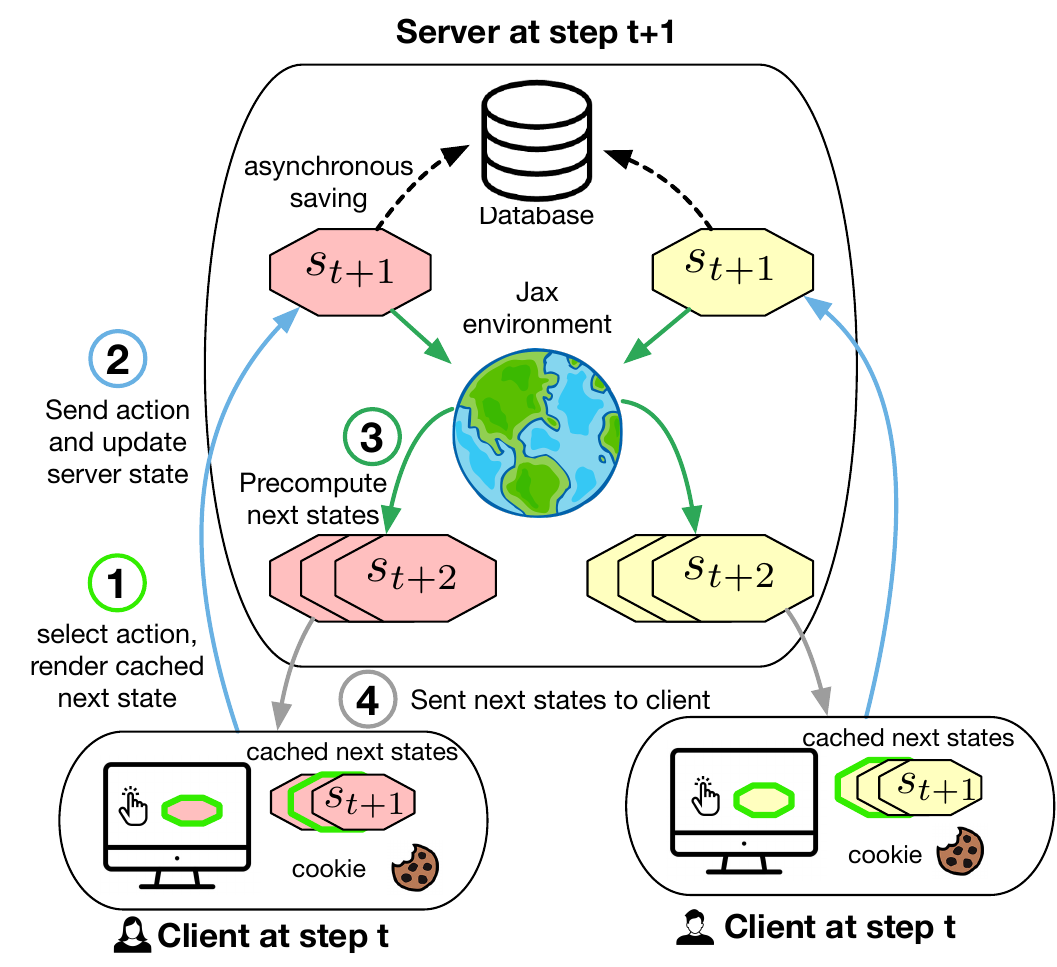}
    \vspace{-15pt}
    \caption{\textbf{How NiceWebRL leverages Jax to enable a single server to interact with multiple users}. Jax's functional paradigm prohibits inter-user state interference since each user has isolated environment states. Jax enables fast parallel computation of future states. Caching these states client side reduces latency.}\label{fig:envstage-explanation}
\end{wrapfigure}
human interaction with Jax-based virtual environments. 
The fundamental structural unit in \nicewebrl~is the stage object, which represents distinct phases of a web experiment.
Stages can display information, collect feedback via forms, or enable environment interaction. 
Researchers define experiments by sequencing different stage types. 
There are three main stage types: instruction stages, feedback stages, and environment interaction stages.
For example, a researcher could have an instruction stage, followed by training and evaluation environment interaction stages.
We describe our setup more formally in  \S\ref{appendix:formal} and provide more details on and examples of stage types in \S\ref{appendix:stage-types}.

\textbf{Interacting with arbitrary environments on a remote server}. 
The key innovation of \nicewebrl~is the \envstage~object, which takes a Jax-based environment as input and allows participants to interact with the environment until some criteria is met (e.g. a minimum number of successful episodes).
To obtain different environment behaviors across different parts of an experiment, \envstage~objects take in ``environment parameters'' (e.g. a fixed task distribution or friction coefficient) to define different dynamics based on a user's experiment stage.
\envstage~leverages a functional paradigm to decouple (1) the function that defines how the environment will reset and evolve from (2) information about the participant's state and from the environment parameters.
This design allows a single compiled environment to be used across different experiment stages by different participants with their own independent environment state. 
We present an overview in~\figref{fig:envstage-explanation}.
  
\textbf{Reducing latency when presenting observations generated by a remote server.} 
\nicewebrl~precomputes potential next states to reduce latency. When a stage initializes interaction with a participant ($t=0$), the server computes the initial state $s_0$ and observation $o_0$, then immediately computes all possible next states $\{s_1\}$ and observations $\{o_1\}$ for each potential action $a$.
Both the initial observation and all potential next observations are sent to the client before any user interaction occurs.
When the observation renders, time $t_1$ is recorded. When the participant selects an action $a_0$, time $t_2$ is recorded. The client immediately renders the corresponding precomputed observation $o_{1}$ and sends $(a_0, t_1, t_2)$ to the server.
The server then selects the appropriate state from its cache, computes the next set of potential states and observations for all possible actions, and sends these to the client for the next interaction.
This parallel computation of all potential next states and observations, enabled by JAX, reduces network latency and helps provide immediate visual feedback to participants. We summarize this server-client Human-environment interaction protocol in Algorithm~\ref{alg:server_client}.

\textbf{Data management and persistence.} Having a persistent participant state across page reloads or web socket connection issues is key to a fluid experiment. We maintain this in two ways. First, we leverage NiceGUI to track unique identifier information held in browser session cookies. To track a participant's experiment progress, we exploit Jax's functional paradigm to track a user's environment state and current random number generator. We serialize these objects and store them for every environment-interaction in a SQL database. Whenever a connection is reset, we identify a user and reload their information for fluid re-engagement with the experiment.

\textbf{Reducing latency when serving multiple clients}. When serving multiple clients, performing I/O operations for persistence can be resource intensive and block other operations. To mitigate this, we save data asynchronously with a queue-based strategy that leverages stochastic exponential backoffs whenever a save fails. Resaving at exponentially increasing time-periods (with some noise) helps prevent collisions if multiple participants try to save concurrently. This is important for having a responsive UI when the server experiences many parallel participants.

\textbf{Human-AI coordination.} When a human coordinates with an artificial agent, we can have the agent be a part of the environment. When the environment steps, it not only computes the state and observation, it also computes the action that the artificial agents would compute for that state and uses that to predict all possible next states for the participant. Thanks to Jax, environments and learning-based agents can be compiled into one function, reducing latency from this extra computation.

\textbf{AI-assisted task completion.} This is a single-agent setting where an LLM has access to either the environment state $s_t$ or environment observation $o_t$. Two options for leveraging LLMs exist. One option is to interact with the LLM via API calls. A second option is to use a local LLM.
        We provide examples of each in our examples folder.